\documentclass[10pt,twocolumn,letterpaper]{article}

\usepackage{iccv}
\usepackage{times}
\usepackage{epsfig}
\usepackage{graphicx}
\usepackage{amsmath}
\usepackage{amssymb}

\usepackage{bbm}
\usepackage{bm}
\usepackage{float}
\usepackage{epstopdf}

\usepackage[pagebackref=true,breaklinks=true,letterpaper=true,colorlinks,bookmarks=false]{hyperref}

\iccvfinalcopy 

\def\iccvPaperID{224} 

\ificcvfinal\fi
\begin{document}

\title{An Attention-Driven Approach of No-Reference Image Quality Assessment}

\author{
Diqi Chen$^{1,3}$, Yizhou Wang$^{1}$, Tianfu Wu$^{2}$, Wen Gao$^{1}$\\
$^1$Nat'l Engineering Laboratory for Video Technology,\\ Key Laboratory of Machine Perception (MoE),\\
	 Sch'l of EECS, Peking University, Beijing, 100871, China\\
	$^2$Department of ECE and the Visual Narrative Cluster,\\ North Carolina State University \\
	$^3$Key Laboratory of Intelligent Information Processing of Chinese Academy of Sciences (CAS),\\
	 Institute of Computing Technology, CAS, Beijing, 100190, China\\
{\tt\small  \{cdq, Yizhou.Wang, wgao\} @pku.edu.cn, tfwu@ncsu.edu}
}

\maketitle

\begin{abstract}
   In this paper, we present a novel method of no-reference image quality assessment (NR-IQA), which is to predict the perceptual quality score of a given image without using any reference image. The proposed method harnesses three functions (i) the visual attention mechanism, which affects many aspects of visual perception including image quality assessment, however, is overlooked in the NR-IQA literature. The method assumes that the fixation areas on an image contain key information to the process of IQA. (ii) the robust averaging strategy, which is a means \--- supported by psychology studies \--- to integrating multiple/step-wise evidence to make a final perceptual judgment. (iii) the multi-task learning, which is believed to be an effectual means to shape representation learning and could result in a more generalized model.
   To exploit the synergy of the three, we consider the NR-IQA as a dynamic perception process, in which the model samples a sequence of ``informative'' areas and aggregates the information to learn a representation for the tasks of jointly predicting the image quality score and the distortion type.
   The model learning is implemented by a reinforcement strategy, in which the rewards of both tasks guide the learning of the optimal sampling policy to acquire the ``task-informative" image regions so that the predictions can be made accurately and efficiently (in terms of the sampling steps). The reinforcement learning is realized by a deep network with the policy gradient method and trained through back-propagation.
   In experiments, the model is tested on the TID2008 dataset and it outperforms several state-of-the-art methods. Furthermore, the model is very efficient in the sense that a small number of fixations are used in NR-IQA.
\end{abstract}

\section{Introduction}
\begin{figure*}
\begin{center}
\includegraphics[width=0.8\linewidth]{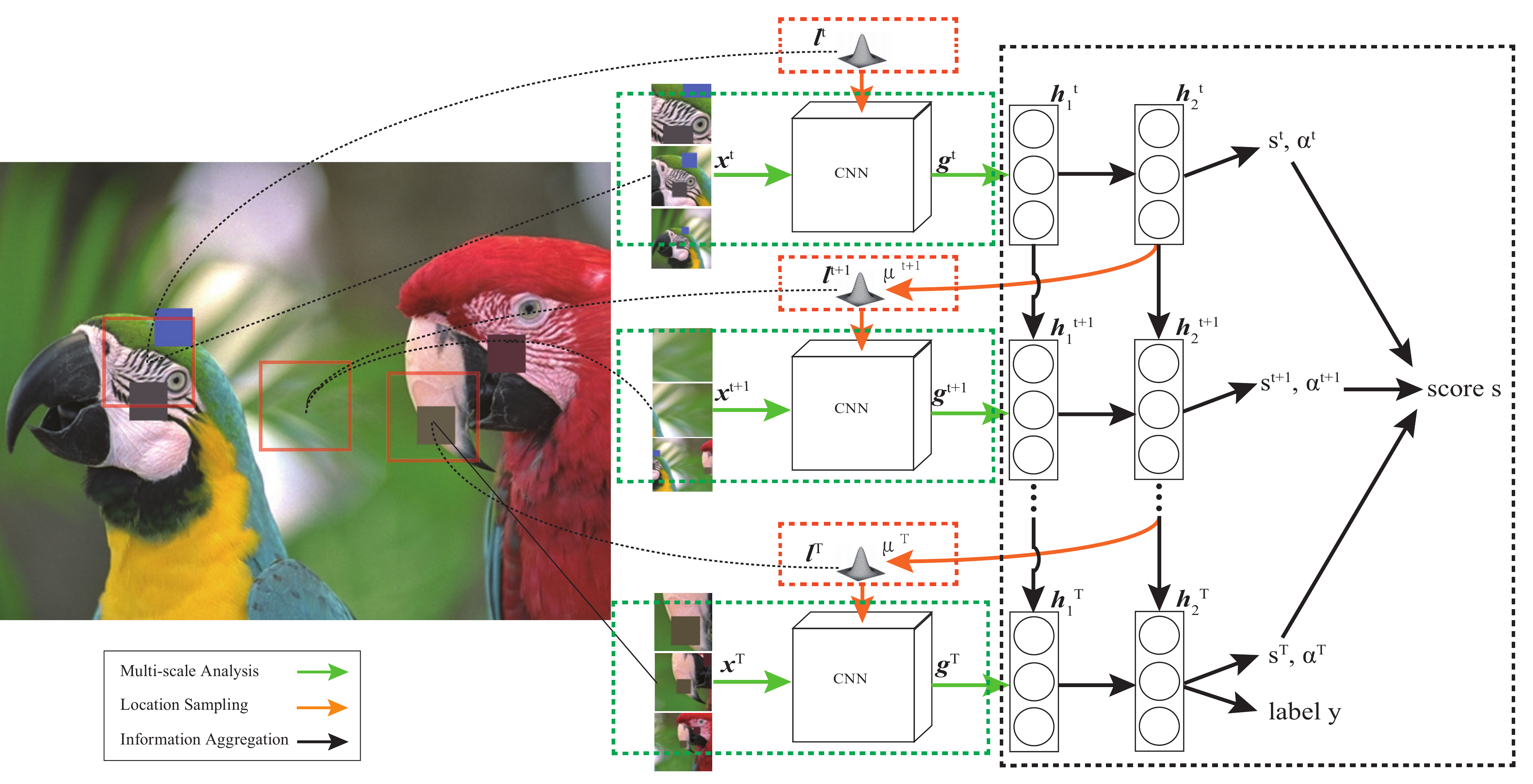}
\end{center}
  \caption{The overall architecture of the proposed model for NR-IQA. \textit{Left}: An image with the local block-wise distortion (the ``block masks") on the two foreground objects. Three fixation areas are illustrated by the orange squares. Centered at each fixation position, three multi-scale patches are extracted and then normalized to the same size to simulate the foveal vision. \textit{Right}: Illustration of the proposed network composed of three components: the multi-scale image analysis module (the green boxes) \--- weights shared CNN used to extract the features to update the recurrent layer, the location sampling module (the orange boxes) \--- a stochastic node learns to select where to attend to next, and the information aggregation module (the black box) \--- a recurrent neural network (RNN) aggregates information along the attentional saccadic path for multi-task learning.
}
\label{model}
\end{figure*}

In the era of big data, an enormous amount of visual data is making its way to end consumers through mobile devices, social media, HDTV, etc. Since the applications are so broad and diverse, it becomes increasingly important to improve the quality of experience for consumers. Automatic IQA becomes an indispensable module of a service system, such that it is able to tell the perceptual quality of its content (here images) and then optimizes the delivered services accordingly.

There are two main schema in IQA: full-reference (FR) IQA~\cite{SSIM, zhang2014vsi, zhang2012sr} and no-reference (NR) IQA~\cite{tang2011learning, hou2015blind, li2015no, kang2014convolutional, kang2015simultaneous, tang2014blind, hou2015saliency, zhang2015som}. The former requires a ``clean", pristine reference image with respect to which the quality of the distorted image is assessed, and the latter takes only the distorted image to be assessed as input and thus is more applicable. This paper focuses on NR-IQA.

The challenges in NR-IQA include many factors: the ``what" issue \--- unknown types of distortion (some are local, e.g., local image regions are distorted; some are global, e.g., pervasive additive noise contaminating all the pixels in an image),
the ``where'' issue \--- unknown spatial distribution of distortions, (e.g., where the degraded regions locate in an image)
and the ``how" issue \--- unknown mechanism about how to aggregate all the information collected from the distorted regions as well as from other regions of an image for quality assessment.

In this paper, we address the above issues by presenting a novel method of NR-IQA. The proposed method is inspired by the following three streams of studies.\\
(i) The visual attention mechanism, which affects many aspects of visual perception including image quality assessment. By observing how human subjects assess the quality of images, we assume that eye fixation areas on an image contain ``key information" for IQA. However, in the IQA literature there is little studies about how to organically integrate the attention mechanism into IQA, e.g. learning the strategy about where to attend on an image that related to the IQA task.\\
(ii) The robust averaging strategy~\cite{de2011robust}, which is a computational mechanism of perceptual judgment \--- supported by psychology studies \--- that integrates multiple/step-wise evidence to make a final judgment. By adopting this strategy, the final image quality score is the weighted average of the scores predicted from a number of attended areas in an image.\\
(iii) The multi-task learning, which is believed to be an effectual means to shape representation learning and could result in a more generalized model. Here, besides predicting image scores, we empower the model to classify distortion types of an image to be assessed.

To exploit the synergy of the three, we consider the NR-IQA as a dynamic perception process, in which the model samples a sequence of ``informative'' areas and aggregates the information to learn a representation for the tasks of jointly predicting the image quality score and the distortion type. Figure~\ref{model} illustrates the proposed model. It is composed of three components:
\begin{itemize}
\item The multi-scale image analysis module: it is implemented by a weight-sharing-CNN (the green boxes in Figure~\ref{model}), which extracts multi-scale image features around a fixation point. We extract three image patches of different scales centered at a fixation point. The CNN learns the feature representation for quality assessment. This component aims at solving the ``what'' issue mentioned above in an end-to-end learning fashion.
\item The location sampling module: it is implemented by a stochastic node (the orange boxes in Figure~\ref{model}), which learns to select ``where" to attend to next based on the integrated information about what the model has seen so far. It predicts the IQA-task-related regions such that the next selected fixation will be sufficiently informative.

\item The information aggregation module: it is implemented by an RNN (the black box in Figure~\ref{model}), which aggregates information along a saccadic path to compute the final predictions, i.e., the image quality score and the distortion type. It captures both local information and global information in the sequential unfolding.  It learns to resolve the ``how'' issue in NR-IQA stated above. The representation is shaped through multi-task learning. Inspired by {\em the robust averaging strategy} for perceptual judgment~\cite{de2011robust} (which takes both the ``strength'' and ``reliability'' of evidence into consideration when making a final perceptual judgment), our model predicts the final score as the weighted averaged of the scores predicted at the attended areas. The weights of the scores are learned to signify the ``reliability" of the score prediction at the attended regions.
\end{itemize}

Inspired by~\cite{williams1992simple}, the model learning is implemented by a reinforcement strategy. The rewards of both tasks (score prediction and distortion type classification) guide the learning of the optimal location sampling policy to acquire the ``task-informative" image regions so that the predictions can be made accurately and efficiently (in terms of the sampling steps). The reinforcement learning is realized by a deep network with the policy gradient method and trained through back-propagation.

In experiments, the model is tested on the TID2008 dataset~\cite{tid2008} and it outperforms several state-of-the-art methods. Furthermore, the model is very efficient in the sense that a small number of fixations are used in NR-IQA.

\section{Related Work}

We briefly review the application of deep learning models for NR-IQA, the IQA methods using objectness/saliency, and a related attentional model.

{\bf Neural Networks for NR-IQA:} Deep learning provides an approach to learning a mapping from raw input or low-level features into scores of image perceptual quality. These methods avoid delicately designing hand-crafted features. Kang \etal~\cite{kang2014convolutional} propose a patch-based NR-IQA method. They first uniformly sample image patches at a predefined scale, then train a CNN to predict a quality score for each image patch and average the scores of the patches as the holistic image score. They further propose a multi-task CNN~\cite{kang2015simultaneous} to classify the distortion type of each patch in addition to the quality score prediction. Our method is quite different from theirs, even in the distortion type classification part, we do not classify the distortion type of each attended patch; instead, we classify a whole image based on the ``aggregated information" over a sequence of attended regions.

In addition to the patch-based methods, some methods combine hand-crafted low-level features with deep networks as an alternative approach. For example, Tang \etal~\cite{tang2014blind} first extract the LBIQ features~\cite{tang2011learning} and feed the features into a Restricted Boilzman Machine to predict image quality scores. Hou \etal~\cite{hou2015blind} pose IQA problem as a classification problem. They slot images into different categories according to the image quality and propose a quality pooling method under the Bayesian framework to predict quality scores.

{\bf Objectness and Saliency in IQA:} Although there lacks of literature that organically fuses the attentional mechanism into the NR-IQA, the semantic objectness or saliency has been applied. Objectness and saliency are static property of image regions, whereas attention is an active perception process of an observer. Liu \etal~\cite{liu2011visual} determine the final score of an image by averaging the predicted patch scores with weights. The patch weights are the saliency values obtained from eye-tracking data. The performance gain of the method justifies the importance of introducing visual attention to IQA. Zhang and Li~\cite{zhang2012sr} argue that visual saliency and perceptual quality are highly related, and they utilize the relationship between a reference image saliency map and its distortion image saliency map to predict image quality scores. Zhang \etal~\cite{zhang2015som} propose an IQA algorithm using object-like regions. They assume that semantic regions contribute to perceptual quality assessment. Hou and Gao~\cite{hou2015saliency} propose a saliency-guided framework whose idea is similar to~\cite{zhang2015som}. In summary, these methods exploit image saliency maps in post-processing, \ie, adopt saliency-weighted average score rather than a uniform average score as final prediction. Zhang \etal~\cite{zhang2016benchmarking} study different combinations of different saliency models and IQA methods.

We also believe that image quality assessment heavily depends on the way how we attend to images. Hence, we explicitly model the attention process and learn the attention policy from data.

{\bf Recurrent Attentional Models:} Recently, deep learning models with attentional mechanism receive a lot of interest. The soft attentional models~\cite{sorokin2015deep, kuen2016recurrent} implement deterministic attention mechanism trained by normal backpropagation.  Kuen \etal~\cite{kuen2016recurrent} realize the attention mechanism through the differentiable spatial transformer~\cite{jaderberg2015spatial} and recurrent connections to refine saliency map step by step. Stochastic attention in the hard attentional models~\cite{mnih2014recurrent, ba2014multiple, sorokin2015deep} are often optimized by the REINFORCE algorithm~\cite{williams1992simple}. Implementing the similar idea, Mnih \etal~\cite{mnih2014recurrent} propose a well-designed attentional model with RNN for object recognition and Ba \etal~\cite{ba2014multiple} recognize and localize multiple objects by maximizing a variational lower bound. Sorokin \etal~\cite{sorokin2015deep} propose a soft attention mechanism designed as element-wise multiplication with importance vectors and a hard attention mechanism optimized by the REINFORCE algorithm.

Compared to the above models, our model also integrates the attentional mechanism but with different ingredients. (i) Our model is multi-task, \ie it jointly optimizes the performance of two closely related tasks to learn representation that leads to a more powerful attention policy. (ii) Consequently, the reward function of the reinforcement learning is enriched with multi-task rewards. Such enriched rewards empower the learned policy being capable of capturing the ``task-informative" regions so that the information are aggregated and predictions are made more accurately and efficiently. (iii) The robust averaging mechanism of perceptual judgment is implemented into the network architecture and learning. (iv) The multi-scale analysis is introduced into the network to emulate the foveal vision and provide contextual information of fixations.

\section{The Proposed Model and Learning}

In this section, we introduce the problem definition, illustrate each component of our model in detail and explain how to jointly learn knowledge about distortion type, perceptual quality, and attention policy.

As shown in Figure~\ref{model}, the proposed model consists of three main parts --- a CNN for multi-scale image feature extraction, a stochastic node for location sampling and recurrent connection for information aggregation. The ultimate goal is to predict the quality score $s$ of an input image $\bm{x}$. Starting from an initial location $\bm{l}^0$, which can be randomly selected in the image during training, at each time $t$, the proposed model learns/extracts features from three normalized multi-resolution patches $\bm{x}^t$ clipped from $\bm{l}^{t}$ and updates the two recurrent layers $\bm{h}_1^t$ and $\bm{h}_2^t$. Based on $\bm{h}_2^t$, our model predicts the next location $\bm{l}^{t+1}$. The model also predicts the image quality scale $s^t$ and the weight $\alpha^t$ signifying reliability of the score prediction. Repeat this procedure for $T$ steps, and we obtain a sequence of locations $\bm{l}=\{\bm{l}^i\}_{i=0}^{T-1}$ and the information on each location is aggregated into $\bm{h}_2^T$. Label $y$ which denotes distortion type of $\bm{x}$ is predicted based on $\bm{h}_2^T$ as an auxiliary task. Then the final quality score of the input image is computed by $s=\sum_{t=t_0}^Ts^t\alpha^t$.

\subsection{The Model Components}

{\bf The Multi-Scale Image Analysis Module:} This module learns a multi-scale representation of an attended region. At step $t$, the output $\bm{g}^t = f_g(\bm{x}, \bm{l}^{t}; \bm{\theta}_g)$, where $\bm{\theta}_g$ is the parameter.

We use multi-resolution images to emulate the foveal vision of human eyes. The fovea is at the center of the retina, where visual signals are captured with high-resolution and processed with details. Regions outside the fovea are peripheral regions, which perceive visual patterns with less details and the degradation grows with eccentricity. Human beings move and fixate their eyes at the task-informative areas with the aid of the peripheral vision so that they are able to acquire/process task-related information efficiently~\cite{freeman2011metamers}. Here, we extract three patches at different scales centered at the same fixation point and normalize them into $32\times32$ patches. These multi-scale patches emulate the foveal and peripheral signals of an attended area. The normalized patches are stacked together and fed to the CNN.

We adopt the multi-scale convolution kernels as in ~\cite{szegedy2015going} to make the computation efficient. We also treat the sampled fixation location as a feature and feed it into a fully connected layer of the CNN. Our model concatenates two hidden layers of $\bm{x}^t$ and $\bm{l}^t$ in the CNN and connects them to another fully connected layer, then outputs $\bm{g}^t$.

{\bf The Location Sampling Module: } This module samples the locations of the attention areas in an image. The output $\bm{l}^t$ is influenced by the hidden state $\bm{h}_2^t$ from the last recurrent layer and parameters $\bm{\theta}_l$: $\bm{l}^{t+1} = f_l(\bm{h}_2^t; \bm{\theta}_l)$. We assume that each dimension of the next location follows a Gaussian distribution independently with the same fixed standard deviation.

The locations are sampled stochastically in the training stage and use the mean of the Gaussian during testing. The stochastic sampling is a common strategy to enable the exploration in reinforcement learning. Firstly, we predict the mean of the Gaussian distribution by $\bm{\mu}^{t+1} = \phi(\bm{W}_{rl}\bm{h}_2^t + \bm{b}_l)$.  $\phi$ is the HardTanh activation function limiting $\bm{\mu}^{t+1}$ into appropriate range ($[-1, 1]$ in this work). Then the next attention location is sampled from a Gaussian distribution $\bm{a}^{t+1} \sim p(\cdot | \bm{\mu}^{t+1}, \sigma)$, where $\sigma$ is the standard deviation for the x-y dimension of the location.

We learn location sampling policy by reinforcement learning guided by the enriched multi-task rewards, so that the model is able to sample a sequence of ``informative" areas and aggregates the information to jointly predicting the image quality score and the distortion type. The details of the learning will be discussed in \ref{sec:learning}.

{\bf The Information Aggregation Module: } The RNN is adopted to learn the internal mechanism of information aggregation across the fixation areas.

A human expert judges the perceptual quality of images after scanning a sequence of attended areas. We employ a two-layer RNN to aggregate information at each time step, then predicts the distortion type and image quality score. At the same time, the model also predicts the next fixation location $\bm{l}^{t}$ at time $t$. The first recurrent layer is computed as
\begin{equation} \label{test}
\bm{h}_1^t = \varphi(\bm{W}_{gh}\bm{g}^t + \bm{W}_{hh}\bm{h}_1^{t-1}+\bm{b}_h),
\end{equation}
where $\bm{W}_{gh}$ denotes the connection weights from $\bm{g}^t$ to the hidden layer $\bm{h}$ and $\bm{W}_{hh}$ denotes the connection of the hidden layer to itself, $\bm{b}_h$ is the bias and $\varphi$ is ReLU activation function. We adopt the same way for calculation of $\bm{h}_2^t$ based on $\bm{h}_1^t$ as input.

The distortion classification and the quality prediction are two different tasks. Although predicting quality score is our ultimate goal, we observe that the classification task not only helps to generalize the learning but also enrich the reward for learning the attention policy. The model predicts the distortion type $\hat{y}$ through two fully connected layers and a softmax layer in the RNN.

In quality score prediction, we adopt the perceptual judgement mechanism of robust averaging~\cite{de2011robust}. According to ~\cite{de2011robust}, an optimal agent will make judgments based on the strength and reliability of decision-relevant evidence. A plausible computational mechanism of the perceptual judgement can be a multi-element averaging model, where the weights of the variables/strengths correspond to the reliability of the evidence. In NR-IQA context, the decision-relevant evidence are the attention image regions, the strengths of the evidence are the predicted quality scores, the reliability is the weights $\{\alpha^t\}$ of the linear averaging model. The weights measure the reliability of the score prediction, and they are learned to optimize the overall reward function of the score prediction and distortion classification.

The unnormalized weight at time $t$ is estimated by $\alpha^t=Linear(\bm{h}_2^T)$ and the predicted score is $s^t = \varphi(Linear(\varphi(Linear(\bm{h}_2^T))))$. We use a softmax layer to normalize the weights  $\{\alpha^t\}$ to make them sum to one. The final score is predicted by $s=\sum_{t=t_0}^{T}\alpha^t s^t$.

\subsection{Learning}
\label{sec:learning}
There are three terms in the final loss function $L = L_{cla} + \lambda L_{reg} - \alpha J_{rein}$, where $\lambda$ and $\alpha$ are free parameters, $L_{cla}$ is the $\log$ softmax loss of the distortion classification and $L_{reg}$ is the mean average error of the quality score prediction.
\begin{figure}
\begin{center}
\includegraphics[width=0.8\linewidth]{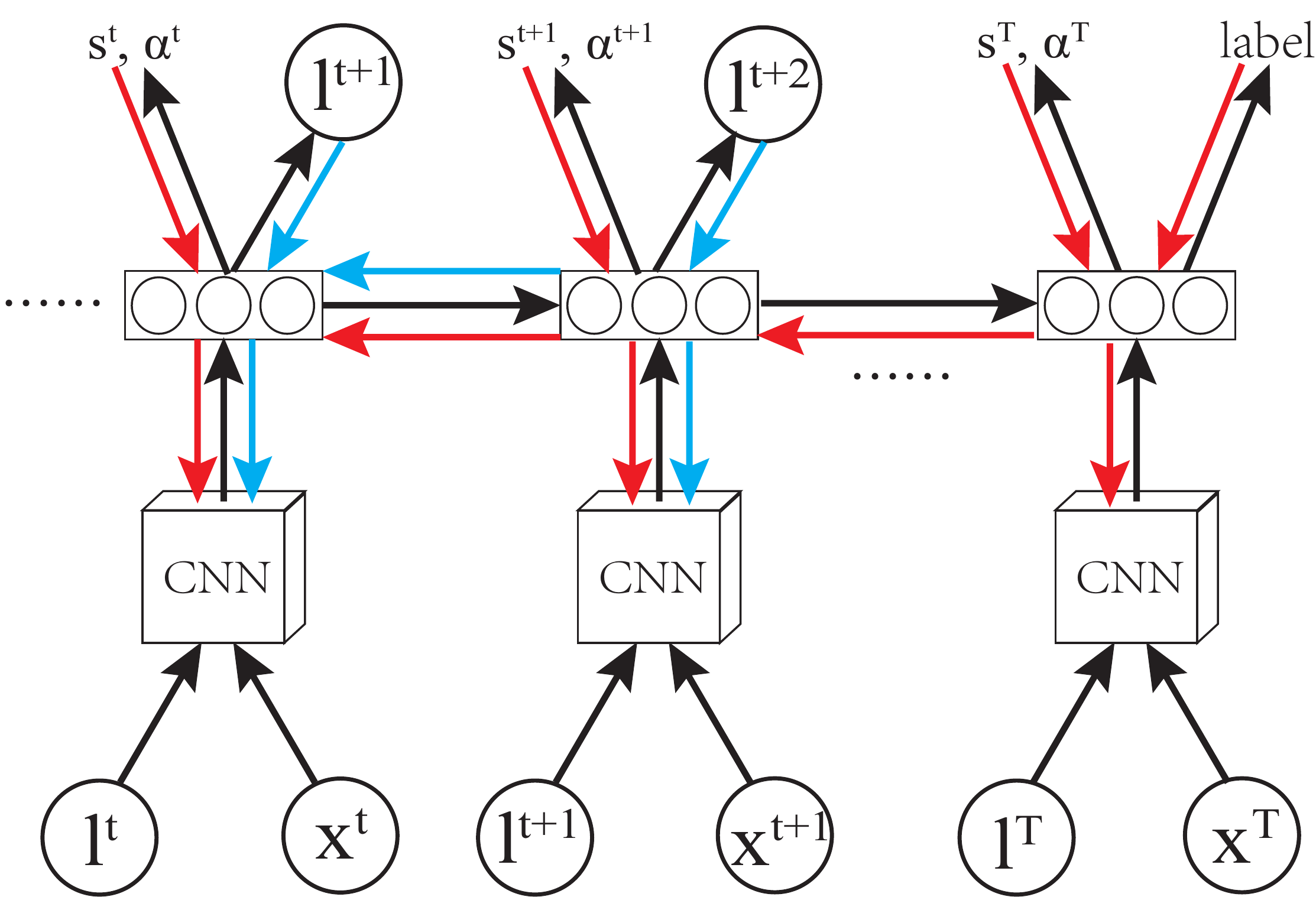}
\end{center}
\caption{Illustration of the backpropagation gradients flow, the red arrows denote the gradients based on supervised signals and the blue arrows denote the gradients of the REINFORCE algorithm.}
\label{fig:BP}
\end{figure}
$J_{rein}$ is the reinforcement learning term which is the expectation of accumulated reward. In the perspective of reinforcement learning, the hidden layer $\bm{h}_2^t$ in our framework represents the states, location prediction is the action, and the predicted Gaussian distribution represents the policy. 
We define the reward function as
\begin{equation} \label{reward}
R = \sum_{t=1}^{T}r^t = r^T = \left\{
\begin{array}{rcl}
1       &      & y=y\prime\ \ \rm{or}\ \ \|s - s\prime\| < \sigma\\
0     &      & \rm{otherwise}
\end{array} \right.
\end{equation}
where $\sigma$ is a threshold to control the policy for assigning reward to the score prediction task, and $y\prime$ is the groundtruth distortion type and $s\prime$ is the groundtruth quality score. Empirically, we set $\sigma=0.7$. The reward equals to $1$ when the classification is correct or the score prediction is accurate enough; 0 otherwise. Because our model only makes predictions at the final step, the cumulative reward $R$ is actually the reward just for the final step $r^T$. Therefore, the goal of the learning is to classify distortion and predict score accurately. 

$J_{rein}$ is approximated by
\begin{equation} \label{reinfoce_loss}
\nabla_{\bm{\theta}_R} J_{rein} \approx \frac{1}{M} \sum_{j=1}^{M} \sum_{t=1}^{T} \nabla_{\bm{\theta}_R} \log p(\bm{a}_{j}^{t} | \bm{\mu}_{j}^{t}, \sigma)R_{j},
\end{equation}
where $j$ is index of the training images and $M$ is the number of images. Intuitively, we learn the sampling policy of the selective attention mechanism by maximize the above $\log$ likelihood function guided by the reward. 
In our case, $\bm{a}_{j}^{t}$ follows the Gaussian distribution parameterized by $\bm{\mu}_{j}^{t}$ and $\sigma$, so the derivative of $J_{rein}$ w.r.t. $\bm{a}_{j}^{t}$ is
\begin{equation} \label{derivative}
\frac{\partial J_{rein}}{\partial \bm{a}_{j}^{t}} = R_{j} \frac{\partial \log p(\bm{a}_{j}^{t} | \bm{\mu}_{j}^{t}, \sigma)}{\partial \bm{a}_{j}^{t}} = - \frac{R_j}{\sigma^2}(\bm{a}_{j}^{t} - \bm{\mu}_{j}^{t}),
\end{equation}
which indicates that the proposed model tends to learn the mean of Gaussian as the center of the informative attended locations.

The model is trained with the Back-Propagation Through Time (BPTT) algorithm. As shown in Figure~\ref{fig:BP}, the black arrows denote the forward computation flow, the red arrows represent the backpropagation flow based on the supervised loss and the blue arrows represent the back-propagation flow of the reinforcement learning loss.

\begin{table*}
\begin{center}
\begin{tabular}{|l|l|l|l|}
\hline
 & \multicolumn{3}{c|}{FR-IQA} \\ \hline
Method & PSNR & SSIM\cite{SSIM} & VSI\cite{zhang2014vsi} \\ \hline
SROCC & 0.749 & 0.814 & \textbf{0.924} \\ \hline
LCC & 0.738 & 0.821 & \textbf{0.902} \\ \hline
\end{tabular}
\begin{tabular}{|l|l|l|l|l|l|l|l|l|}
\hline & \multicolumn{8}{c|}{NR-IQA} \\ \hline
Method & CNN\cite{kang2014convolutional} & CNN++\cite{kang2015simultaneous} & Tang \etal~\cite{tang2014blind} & \textbf{CNN\_MT} & \textbf{CNN\_MT+S} & \textbf{RL} & \textbf{RL+M} & \textbf{RL+M+R} \\ \hline
SROCC & 0.548 & 0.633 & \textbf{0.841} & 0.816 & 0.825 & 0.646 & 0.819 & 0.833 \\ \hline
LCC & 0.614 & 0.675 & - & 0.821 & 0.832 & 0.701 & 0.824 & \textbf{0.841} \\ \hline
\end{tabular}
\end{center}
\caption{Evaluation on TID2008.}
\label{TID2008results}
\end{table*}

\subsection{Implementation Details}
To preprocess the images, we first turn the RGB images into gray scale images, then apply a local contrast normalization method on them.

In the multi-scale image analysis module, the patches sizes at three scales are $32\times32$, $96\times 96$ and $288\times 288$. All the patches are normalized to $32\times32$. We use four multi-scale convolution layers with $3\times3$, $5\times5$ and $1\times1$ convolution kernels. The ratio of numbers of the three types of kernels is $2:1:1$, and the numbers of the kernels in layers of CNN are $32,64,64,128$. The spatial pooling size is $8\times8$ in the last convolution layer and $2\times2$ in the first and the third convolution layers. The two hidden layers of RNN for both tasks have 256 neurons and all of the other hidden layers have 128 neurons. We use ReLU for all the convolution and the linear layers in the multi-scale image analysis module and the RNN.

We use an adaptive gradient descent methods Adam~\cite{dozatincorporating} with momentum $0.9$ as our optimization method. In the loss function, parameter $\lambda$ for the score prediction task is set to $1$ and $\alpha$ for the reinforcement loss is set to $0.01$. The initial learning rate is $0.001$, and we train the model with $1000$ epochs and linearly decay the learning rate to $0.0001$. To encourage the exploration of the location sampling policy, the standard deviation $\sigma$ of the Gaussian distribution is linearly declined from $0.16$ to $0.1$ after training for $100$ epoches. We apply the $\epsilon$-greedy method for location sampling and $\epsilon$ is linearly declined from $0.1$ to zero after training for $100$ epoches. The number of sampled locations $T$ is set to be five. We find that sometimes the overflow of locations is serious, at the beginning of training. We apply a small trick to make learning more stable at the very start. If we detect the sum of $\bm{\mu}^t$ which is the mean of Gaussian is larger than a threshold, we randomly reset the parameters in the location sampling module.

\begin{figure}
\begin{center}
\includegraphics[width=1\linewidth]{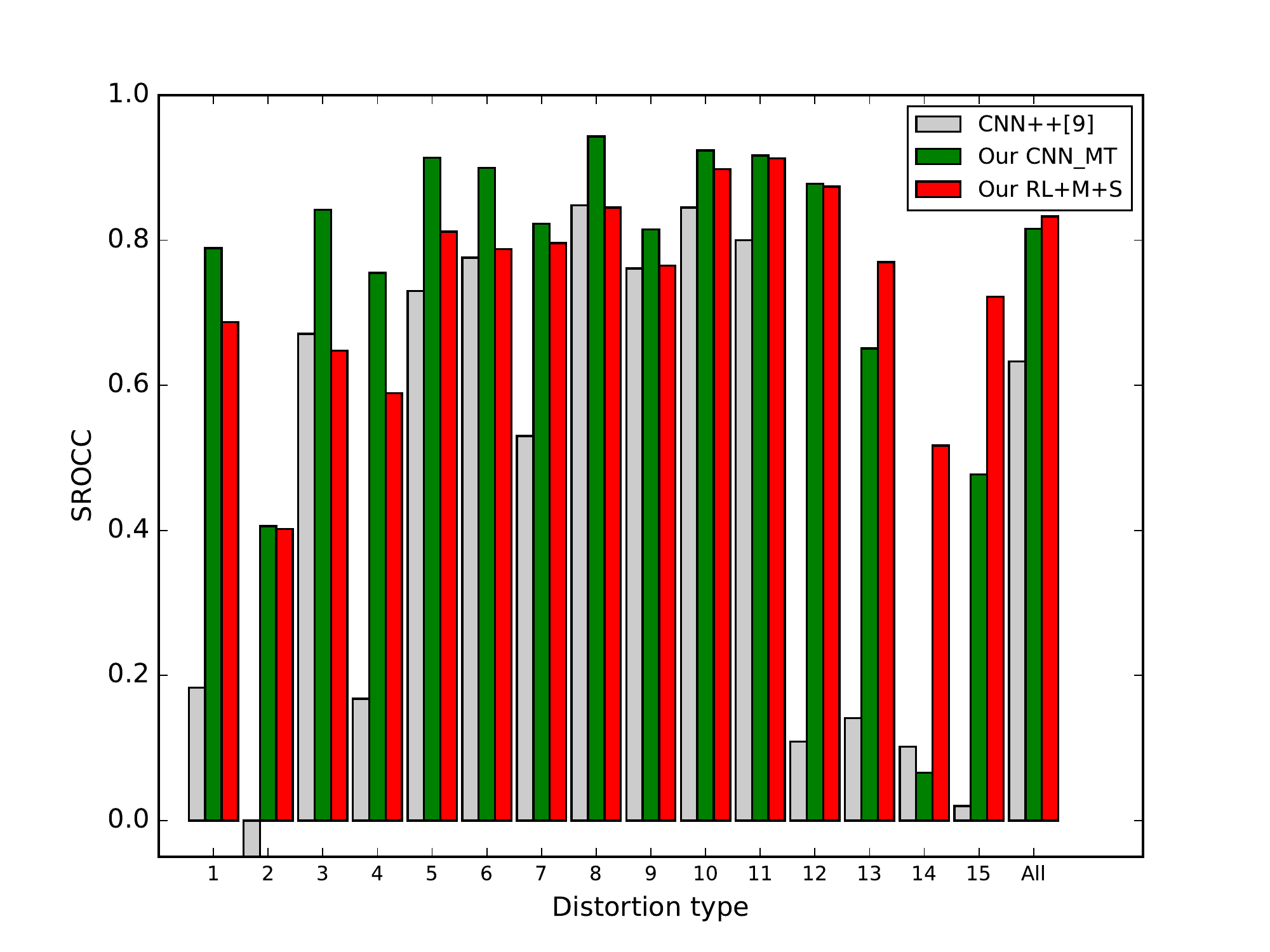}
\end{center}
   \caption{SROCC on TID2008 for each distortion type.}
\label{fig:TID2008results2}
\end{figure}

\section{Experimental Results}

\subsection{Experiment Settings}
TID2008~\cite{tid2008}: This dataset consists of $25$ reference images, $17$ types of distortions and four levels of each type of distortion. There are in total $1700$ distorted images, each of which is labeled with a Mean Opinion Score (MOS) between $0$ and $9$.

Evaluation: We choose the Pearson linear correlation coefficient (LCC) to measure the prediction accuracy and the Spearman rank order correlation coefficient (SROCC) to measure the prediction monotonicity.

Local contrast normalization used in our method is not applicable to the ``mean shift" and ``contrast change" distortions, so them are neglected in our experiments. We ignore the last reference image because it is not a natural image.

We select $60\%$ of the reference images and the associated distorted images as the training set, $20\%$ and the rest $20\%$ as the validation set and the testing set, respectively. The results are reported based on median of five random splits. During testing, we set the initial location to be the center and choose the model parameters with the highest SROCC in the validation.

\begin{table}
\begin{center}
\begin{tabular}{|l|l|l|l|l|l|}
\hline
SROCC   & 12    & 13    & 14     & 15    & Overall \\ \hline
CNN\_MT & \textbf{0.752} & 0.738 & \textbf{0.111}  & 0.630 & 0.617   \\ \hline
RL+M+R  & 0.683 & \textbf{0.748} & 0.106  & \textbf{0.739} & \textbf{0.691}   \\ \hline
\end{tabular}
\begin{tabular}{|l|l|l|l|l|l|}
\hline
LCC   & 12    & 13    & 14     & 15    & Overall   \\ \hline
CNN\_MT & \textbf{0.702} & 0.683 & 0.065  & 0.383 & 0.613   \\ \hline
RL+M+R  & 0.686  &  \textbf{0.707} & \textbf{0.071} & \textbf{0.740} & \textbf{0.658}   \\ \hline
\end{tabular}
\end{center}
\caption{Results of local distortion types on TID2008.}
\label{local_distortion}
\end{table}

\begin{table}
\begin{center}
\begin{tabular}{|l|l|l|l|l|}
\hline
                    &  SROCC   & Class. Acc. \\ \hline
RL+M+R without multi-resolution &  0.774   & 80.7\%       \\ \hline
RL+M+R                 &  \textbf{0.833}  & \textbf{87.7\%}  \\ \hline
\end{tabular}
\end{center}
\caption{Comparison between with and without multi-resolution context information on TID2008.}
\label{context}
\end{table}

\begin{figure*}[!ht]
\begin{center}
\includegraphics[width=0.9\linewidth]{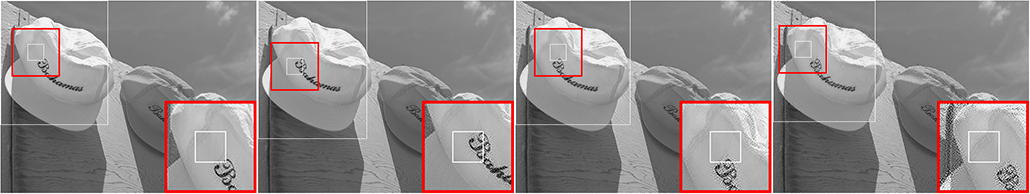}
\end{center}
   \caption{The last attended regions of four images with the masked noise of different levels. Our model locates on or near the most salient region around the black letters.}
\label{fig:attention4}
\end{figure*}

\begin{figure*}[!ht]
\begin{center}
\includegraphics[width=0.9\linewidth]{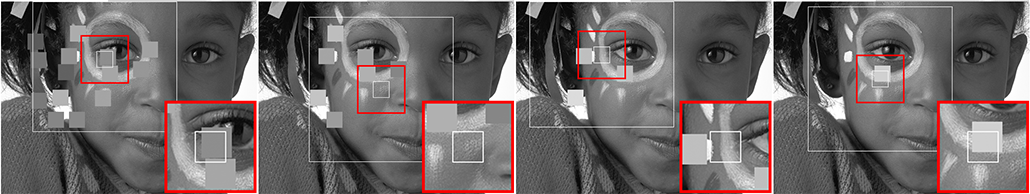}
\end{center}
   \caption{The last attended regions of four images with the local block-wise distortions of different levels. Our model locates on or near the block masks.}
\label{fig:attention15}
\end{figure*}

\begin{figure*}[!ht]
\begin{center}
\includegraphics[width=0.9\linewidth]{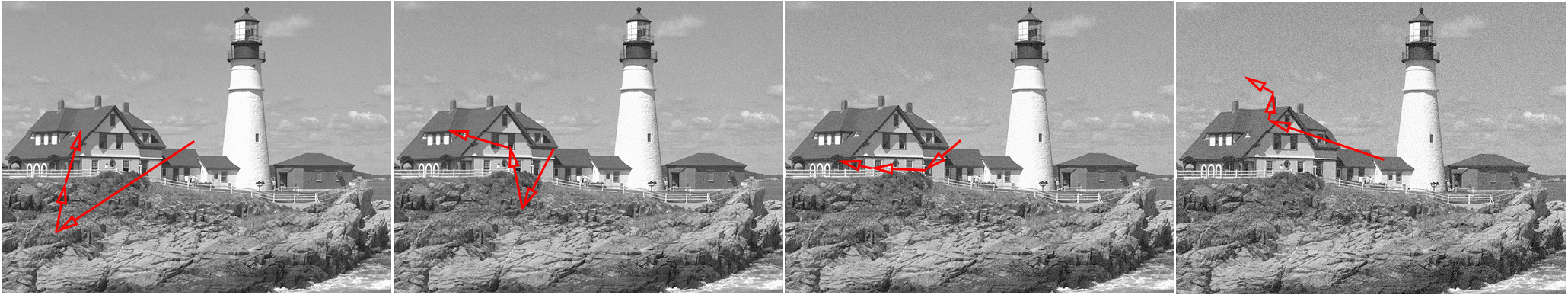}
\end{center}
   \caption{The sampled locations of four images with the high frequency noise of different levels. The fixation moves out into background only in the image with most serious degradation.}
\label{fig:attention5}
\end{figure*}

\subsection{Evaluation on TID2008}
We train the model with the images of the $15$ distortion types together. The overall results are presented in Table~\ref{TID2008results} and SROCC evaluation for each specific distortion type is presented in Figure~\ref{fig:TID2008results2}. We compare our reinforcement learning model with multi-task learning and robust averaging (RL+M+R) against some FR-IQA methods (PSNR, SSIM~\cite{SSIM}, VSI~\cite{zhang2014vsi}) and some NR-IQA methods (CNN~\cite{kang2014convolutional}, CNN++~\cite{kang2015simultaneous}, Tang \etal's method~\cite{tang2014blind}). 
The CNN~\cite{kang2014convolutional} and CNN++~\cite{kang2015simultaneous} are implemented by ourselves following the original settings strictly. Our model outperforms most of the state-of-the-art NR-IQA and even FR-IQA methods on the TID2008 dataset. Tang \etal's method~\cite{tang2014blind} performs better than ours, but their model is pre-trained on a large-scale external dataset.

In order to demonstrate the benefit of the robust averaging strategy, we implement a RL+M model which uses multi-task learning but without robust averaging strategy. As shown in Table~\ref{TID2008results}, the results of our RL+M+R is better than the results of the RL+M ($0.833$ v.s. $0.819$ of SROCC, $0.841$ v.s. $0.824$ of LCC).

In order to justify the importance of multi-task learning, we implement a RL model without multi-task learning and robust averaging strategy. As shown in Table~\ref{TID2008results}, the results of the RL performs much poorer compared with the RL+M and the RL+M+R (SROCC of the RL is $0.646$ and LCC of the RL is $0.701$, while SROCC/LCC of the RL+M and the RL+M+R are larger than $0.81$).

In order to show that the boosted performance is due to our task-driven attentional mechanism, we implement a multi-task CNN (CNN\_MT) with similar structure to our RNN. The training and testing procedures on the CNN\_MT is the same as that on CNN++~\cite{kang2015simultaneous}. As shown in Table~\ref{TID2008results}, the results of the CNN\_MT is worse than our RL+M and RL+M+S.
Furthermore, we combine a saliency model \cite{wang2010measuring} with the CNN\_MT and name it as CNN\_MT+S. First we apply the saliency method to compute saliency maps of the TID2008 images, then use saliency values as the weights to average the scores predicted by CNN\_MT. The results are shown in Table~\ref{TID2008results}, the CNN\_MT+S is better than the CNN\_MT, but worse than our RL+M+R.



Figure~\ref{fig:TID2008results2} shows the SROCC values of each distortion type of different methods. It can be seen that, our model performs better on the images with local distortions, especially for the Type 14, \ie  non-eccentricity pattern noise and the Type 15, \ie local block-wise distortions of different intensity. The CNN\_MT outperforms our model in a few distortion types but performs worse in the overall result. This may indicate that it is not a good strategy to obtain the quality score of an image by averaging the scores of every patches. Instead, our attention-driven model which only uses ``informative" patches is a better method.

{\bf Experiments on Local Distortion Types:} We train our model on the images of four local distortion types on the TID2008. As shown in Table~\ref{local_distortion}, the results of our RL+M+R are better than the results of CNN\_MT ($0.691$ v.s. $0.617$ of SROCC, $0.658$ v.s. $0.613$ of LCC). 

{\bf Learning without Multi-Resolution Information: } In the proposed method, we extract multi-resolution patches. As a reference, we train a model operating on only one $32\times32$ patch each time. This model is compared with the proposed one in Table~\ref{context}. Both the quality assessment results and distortion type prediction results decline when learning without multi-resolution patches.

\begin{figure}
\begin{center}
\includegraphics[width=1\linewidth]{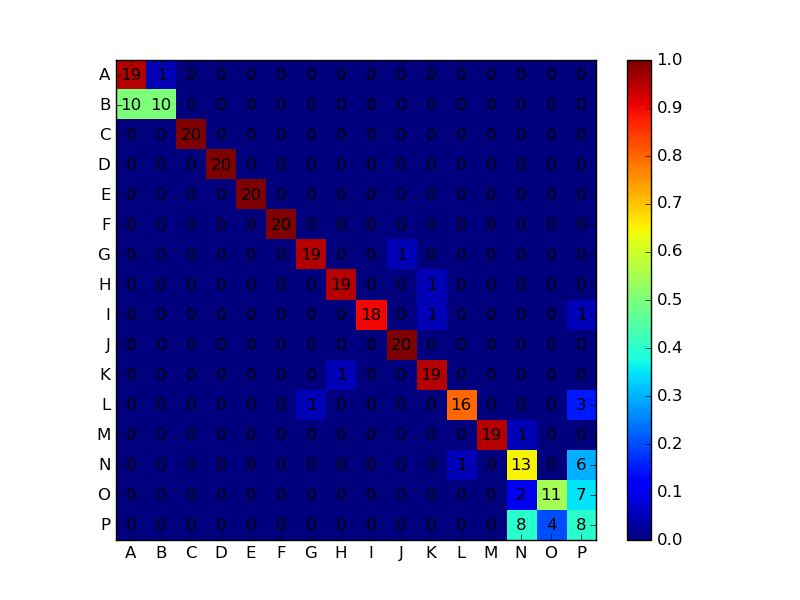}
\end{center}
   \caption{Confusion matrix on testing set}
\label{fig:acc}
\end{figure}

{\bf Classification Task: } On the testing set, our model obtains 87.7\% classification accuracy for the 15 distortion types. The confusion matrix is shown in Figure~\ref{fig:acc}. Half of the images of the Type 2 are misclassified into the Type 1 because they are both additive Gaussian noise, while the Type 2 is operated in the luminance channel, but the Type 1 operates on the color components. The lower right corner of confusion matrix shows that images with very small size of local distortions are hard to be correctly classified.

\subsection{Attentional Locations}
We show some results of the attended patches by our model in Figure~\ref{fig:attention4},~\ref{fig:attention15} and~\ref{fig:attention5}.

In Figure~\ref{fig:attention4} and Figure~\ref{fig:attention15}, we magnify the sampled patches at the bottom right corner of each image. 
Figure~\ref{fig:attention4} shows that the last attended regions of four images with the masked noise. The degradation has different intensity. The masked noise is strong in regions of high spatial frequency. The highest spatial frequency regions are the areas around the letters of the left cap. Our model locates this most salient region for all the four distortion levels.

The local block-wise distortion degrades image quality by adding some annoying blocks with different intensity. Figure~\ref{fig:attention15} shows that our model locates the artifact blocks in the last attended region. Notice that in the last two images, even the distortion of only a few blocks is capture.

Figure~\ref{fig:attention5} displays the attentional scanpaths on an image with different levels of high frequency noises. Notice that the scanpaths are different, which indicates that different level of degradation can affect the attention.



\section{Conclusion}
In the paper we propose an attention-driven model with multi-task learning and robust averaging strategy for general no-reference image quality assessment. We consider the NR-IQA as a dynamic perception process. The model learning is implemented by a reinforcement strategy, in which the rewards of both tasks guide the learning of the optimal sampling policy to acquire the “task-informative” image regions so that the predictions can be made accurately and efficiently. 

{\small
\bibliographystyle{ieee}
\bibliography{egbib}
}

\end{document}